\documentclass[11pt,letterpaper]{article}

\usepackage{rotating}
\usepackage{acl2012}
\usepackage{times}
\usepackage{url}
\usepackage{amsmath}
\usepackage{amsthm}
\usepackage{bbm}
\usepackage{amsfonts}
\usepackage{latexsym}
\usepackage{graphicx}
\usepackage{appendix}
\usepackage{booktabs}
\usepackage{multirow}
\usepackage{rotating}
\usepackage{mathtools}
\usepackage{adjustbox}
\usepackage{array}

\long\def\eat#1{\ignorespaces}

\newcommand{\sss}[1]{{\footnotesize #1}}

\eat{
\usepackage[inline,shortlabels]{enumitem}
\setlist*[enumerate,1]{label=$(\arabic*)$}
\setlist*[itemize,1]{label=$\bullet$}
}

\usepackage{cleveref}
\crefname{section}{Section}{Section}
\Crefname{section}{Section}{Section}
\crefname{figure}{Figure}{}
\crefname{algorithm}{Algorithm}{}

\usepackage{algorithm}
\usepackage[noend]{algpseudocode}

\newcounter{notecounter}
\newcommand{\enotesoff}{\long\gdef\enote##1##2{}}

\enotesoff

\newcounter{highprioritynotecounter}
\newcommand{\ehighprioritynotesoff}{\long\gdef\ehighprioritynote##1##2{}}

\ehighprioritynotesoff

\renewcommand{\algorithmicindent}{9pt}
\algnewcommand{\LineComment}[1]{\State \(\triangleright\) {\small \it #1}}
\algnewcommand{\InlineComment}[1]{\hfill \(\triangleright\) {\small \it #1}}
\algrenewcommand\algorithmicindent{1.0em}%

\newcommand{\software}[1]{{\sc #1}}
\newcommand{\word}[1]{{\em #1}}

\newcommand{\saveforCR}[1]{}

\DeclareMathOperator*{\Expect}{\mathop{\mathbb{E}}}

\newcommand{\CC}{\mathcal{C}_{\vbeta}}

\usepackage{color}
\usepackage{xcolor}
\definecolor{darkgrey}{rgb}{0.2,0.2,0.2}
\definecolor{grey}{rgb}{0.9,0.9,0.9}
\definecolor{darkblue}{rgb}{0.0,0.0,0.5}
\definecolor{darkpurple}{rgb}{0.4,0.0,0.4}
\definecolor{darkred}{rgb}{0.5,0.0,0.0}
\definecolor{darkorange}{rgb}{0.5,0.45,0.4}
\definecolor{darkgreen}{rgb}{0.0,0.5,0.0}
\definecolor{darkyellow}{rgb}{1.0,0.65,0.0}
\definecolor{darkergreen}{rgb}{0.0,0.4,0.0}
\definecolor{lightblue}{rgb}{0.8,0.8,1.0}
\definecolor{lightgreen}{rgb}{0.8,1.0,0.8}
\definecolor{lightred}{rgb}{1.0,0.8,0.8}
\definecolor{lightyellow}{rgb}{1.0,1.0,0.8}
\definecolor{lightorange}{rgb}{1.0,0.9,0.8}
\definecolor{lightgrey}{rgb}{0.96,0.97,0.98}
\definecolor{brilliantlavender}{rgb}{0.96, 0.73, 1.0}
\usepackage{tipa}

\usepackage{soul}
\usepackage{color}
\newcommand{\Note}[3]{\sethlcolor{#2}\hl{[\textbf{#1}: #3]}}
\renewcommand{\Note}[3]{}

\usepackage[small]{caption}

\title{\raisebox{1ex}[0in][0in]{\parbox[b]{\linewidth}{\begin{flushright}\footnotesize
        \textmd{\textsf{\textcolor{gray}{Appeared in TACL 2018 and was presented at ACL 2017 (Vancouver, July).  This
          version was \\ prepared in October 2018 and fixes some minor mistakes .}}}\end{flushright}}}\\ \vspace{-1ex}Joint Semantic Synthesis and Morphological Analysis of the Derived Word}
\author{Ryan Cotterell \\
Department of Computer Science \\
Johns Hopkins University \\
{\tt ryan.cotterell@jhu.edu} \\\And
Hinrich Sch{\"u}tze \\
CIS \\
LMU Munich  \\
{\tt inquiries@cislmu.org}}

\renewcommand{\vec}{\boldsymbol}
\newcommand{\vtheta}{{\vec{\theta}}}
\newcommand{\vbeta}{{\vec{\beta}}}
\newcommand{\veta}{{\vec{\eta}}}
\newcommand{\vomega}{{\vec{\omega}}}

\newcommand{\vf}{{\vec{f}}}
\newcommand{\vg}{{\vec{g}}}
\newcommand{\vh}{{\vec{h}}}

\newcommand{\vv}{v}

\newcommand{\vm}{m}

\begin{document}
\maketitle

\def\figref#1{Figure~\ref{fig:#1}}
\def\figlabel#1{\label{fig:#1}\label{p:#1}}
\def\Tabref#1{Table~\ref{tab:#1}}
\def\tabref#1{Table~\ref{tab:#1}}
\def\tablabel#1{\label{tab:#1}\label{p:#1}}
\def\Secref#1{\S\ref{sec:#1}}
\def\secref#1{\S\ref{sec:#1}}
\def\seclabel#1{\label{sec:#1}\label{p:#1}}
\def\eqref#1{Equation~\ref{eqn:#1}}
\def\eqrefn#1{\ref{eqn:#1}}
\def\eqsref#1#2{Equations~\ref{eqn:#1}/\ref{eqn:#2}}
\def\eqlabel#1{\label{eqn:#1}}
\def\subsp#1{P_{\mbox{{\scriptsize\rm #1}}}}

\def\spacesavingparagraph#1{\paragraph{#1}}

%
%
%


\begin{abstract}
  Much like sentences are composed of words, words themselves are
  composed of smaller units. For example, the English word
  \word{questionably} can be analyzed as
  \word{question}$+$\word{able}$+$\word{ly}. However, this {\em
    structural} decomposition of the word does not directly give us a
  \emph{semantic} representation of the word's meaning.
  Since  morphology obeys the principle of compositionality, the semantics of the word
  can be systematically derived from the meaning of its parts. In this
  work, we propose a novel probabilistic model of word formation that
  captures both the \emph{analysis} of a word $w$ into its constituent
  segments and the \emph{synthesis} of the meaning of $w$
  from the meanings of those segments. Our model jointly learns
  to \emph{segment} words into morphemes and \emph{compose} distributional
  semantic vectors of those morphemes.
  We experiment with the model on English
CELEX data and German DErivBase \cite{zeller2013derivbase} data. We show that jointly modeling semantics increases
both segmentation accuracy and morpheme $F_1$ by between 3\%
and 5\%. Additionally,
we investigate different models of vector composition, showing
that recurrent neural networks yield an improvement
  over simple additive models. Finally, we study
   the degree to which the representations correspond to
  a linguist's notion of morphological productivity.
\end{abstract}

\section{Introduction}\seclabel{introduction}
In most languages, words decompose further into smaller units, termed
morphemes. For example, the English word \word{questionably} can be
analyzed as \word{question}$+$\word{able}$+$\word{ly}. This {\em
  structural} decomposition of the word, however, by itself is not a
\emph{semantic} representation of the word's
meaning;\footnote{There are many different linguistic and
  computational theories for interpreting the structural
  decomposition of a word. For example, \word{un-} often
  signifies negation and its effect on semantics can then be
  modeled by theories based on logic. This work addresses
  the question of structural decomposition and semantic
  synthesis in the general framework of
  distributional semantics.}
we further
require an account of how to synthesize the meaning from the
decomposition. Fortunately, words---just like phrases---to a large extent
obey the principle of compositionality: the semantics of the word can
be systematically derived from the meaning of its
parts.\footnote{Morphological research in theoretical and
  computational linguistics often focuses on 
  noncompositional or less compositional phenomena---simply
  because compositional derivation poses fewer interesting
  research problems. It is also true that---just as many
  frequent multiword units are not completely compositional---many frequent derivations (e.g., \word{refusal},
  \word{fitness})
are not completely compositional. An indication 
that non-lexicalized derivations are usually compositional is
the fact that standard dictionaries like \newcite{noad10} list derivational affixes with
their compositional meaning, without a hedge that they can also
occur as part of
only partially compositional forms.
See also \newcite{haspelmath2013understanding}, \S 5.3.6.}
In this work,
we propose a novel joint probabilistic model of word formation that
captures both \emph{structural decomposition} of a word $w$
into its constituent
segments and 
the \emph{synthesis} of $w$'s \emph{meaning} from the meaning of those segments.

Morphological segmentation is a structured prediction task that seeks
to break a word up into its constituent morphemes. The output
segmentation has been shown to aid a diverse set of applications, such
as automatic speech recognition \cite{afify2006use}, keyword spotting
\cite{narasimhanmorphological}, machine translation
\cite{clifton2011combining} and parsing \cite{seeker2015graph}. In
contrast to much of this prior work, we focus on {\em supervised}
segmentation, i.e., we provide the model with gold segmentations during
training time. Instead of \emph{surface} segmentation, our
model performs \emph{canonical} segmentation \cite{cotterell16insideout,cotterell2016canonical,kann16neural},
i.e., it
allows the induction of
orthographic changes together with the segmentation, which is not
typical. For the example \word{questionably}, our model can restore
the deleted characters \word{le}, yielding the canonical
segments
\word{question}, \word{able} and \word{ly}.
In this work, our primary contribution lies in the integration of
continuous semantic vectors into supervised morphological
segmentation---we present a joint model of morphological analysis and
semantic synthesis at the word-level.

We experimentally
investigate three novel aspects of our model.
\begin{itemize}
\item First,
we show that jointly modeling continuous representations
of the semantics of morphemes and words allows us to improve morphological analysis. On the
English portion of CELEX \cite{baayen1993celex}, we achieve a 5 point
improvement in segmentation accuracy and a 3 point improvement in
morpheme $F_1$. On the German DErivBase dataset we achieve a 3 point
improvement in segmentation accuracy and a 3 point
improvement in morpheme $F_1$. \item Second, we explore improved models of vector
composition for synthesizing word meaning.  We find a recurrent neural
network improves over previously proposed additive models. Moreover,
we find that more syntactically oriented vectors
\cite{levy2014dependency} are better suited for morphology than
bag-of-word (BOW) models. \item Finally, we explore the productivity of
English derivational affixes in the context of distributional
semantics.
\end{itemize}

\section{Derivational Morphology}\seclabel{deriv-morph}
Two important goals of morphology, the linguistic study of the
internal structure of words, are to describe the relation between
different words in the lexicon and to decompose them into {\em
  morphemes}, the smallest linguistic unit bearing meaning.
Morphology can be divided into two types: {\em inflectional} and {\em
  derivational}. Inflectional morphology is the set of processes
through which the word form outwardly displays syntactic information,
e.g., verb tense. It follows that an inflectional affix typically
neither changes the part-of-speech (POS) nor the semantics of the
word.  For example, the English verb \word{to run} takes various
forms: \word{run}, \word{runs}, \word{ran} and \word{running}, all of
which convey ``moving by foot quickly'', but appear in complementary
syntactic contexts.

\enote{hs}{
DO WE NEED THIS

NLP tasks that focus on inflection include morphological
tagging \cite{hajivc2000morphological}, lemmatization
\cite{DBLP:conf/emnlp/0009CFS15}, reinflection \cite{dreyer2008latent,cotterell2016sigmorphon}
and paradigm completion \cite{DBLP:conf/naacl/DurrettD13}. 
It is inflectional morphology that most helps
with syntactic tasks, e.g.,  parsing \cite{BohnetNBFGH13}.
}


Derivation deals with the formation of new
words that have semantic shifts in meaning (often including
POS) and is tightly intertwined with lexical semantics
\cite{light:1996:ACL}. Consider the example of the English noun
\word{discontentedness}, which is derived from the adjective
\word{discontented}. It is true that both words share a close semantic
relationship, but the transformation is clearly more than a simple
inflectional marking of syntax. Indeed, we can go one step further and
define a chain of words \word{content} $\mapsto$ \word{contented}
$\mapsto$ \word{discontented} $\mapsto$ \word{discontentedness}.

In the computational literature, derivational morphology has received
less attention than inflectional. There are, however, two bodies of
work on derivation in computational linguistics. First, there is a
series of papers that explore the relation between lexical semantics
and derivation
\cite{LazaridouMZB13,ZellerPS14,pado15:_measur_seman_conten_to_asses_asymm_deriv,kisselew2015obtaining}. All
of these assume a gold morphological analysis and primarily
focus on the
effect of derivation on distributional semantics. The second body of
work, e.g., the unsupervised morphological segmenter
\software{Morfessor} \cite{creutz2005unsupervised}, does not deal with
semantics and makes {\em no distinction} between inflectional and
derivational
morphology.\footnote{\newcite{narasimhan2015unsupervised}
  also make no distinction between inflectional and
  derivational morphology, but their model is an exception
  in that it includes vector similarity
  as a semantic feature. See \secref{related-work} for discussion.}
  Even though the boundary between
inflectional and derivational morphology is a continuum rather than a
rigid divide \cite{haspelmath2013understanding}, there is still the
clear distinction that derivation changes meaning whereas inflection
does not. Our goal in this paper is to develop an account of how the
meaning of a word form can be computed jointly, combining these two
lines of work.


\spacesavingparagraph{Productivity and Semantic Coherence.}
We highlight two related issues in  derivation that
 motivated the development of our model: productivity and semantic coherence.
Roughly, a {\em productive} affix is one that can
still actively be employed to form new words in a language. For example, the
English nominalizing affix \word{ness}
(\word{red}$\mapsto$\word{red}$+$\word{ness}) can be attached
to just about any adjective, including novel forms. In contrast, the
archaic English nominalizing affix \word{th}
(\word{dear}$\mapsto$\word{dear}$+$\word{th},
\word{heal}$\mapsto$\word{heal}$+$\word{th}, 
\word{steal}$\mapsto$\word{steal}$+$\word{th}) 
does not allow
us to form new
words such as \word{cheapth}. This is a crucial issue in derivational
morphology since we would not in general want to analyze new words as
having been formed from non-productive endings; e.g., we do not want
to analyze \word{hearth} as \word{hear}$+$\word{th} (or \word{wugth}
as \word{wug}$+$\word{th}).  Relations such as those between
\word{heal} and \word{health} are {\em lexicalized} since
they no
longer can be derived by productive processes
\cite{bauer1983english}.

Under a generative treatment \cite{chomsky2014aspects} of morphology,
productivity becomes a central notion since a grammar needs to account
for active word formation processes in the language
\cite{aronoff1976word}.  Defining productivity precisely, however, is
tricky; \newcite{aronoff1976word} writes, {\em ``one of the central
  mysteries of derivational morphology \ldots\ [is that]
  \ldots\ though many things are possible in morphology, some are more
  possible than others.''}  Nevertheless, speakers often have clear
intuitions about which affixes in the language are productive.\footnote{It is also important to distinguish productivity from {\em
    creativity}---a non-rule-governed form of word
  formation \cite{lyons1977semantics}. As an example of creativity,
  consider the creation of portmanteaux, e.g., \word{dramedy} and \word{soundscape}.}

%
%

Related to productivity is the notion of {\em semantic coherence}.
The principle of compositionality \cite{frege1892,heim1998semantics}
applies to interpretation of words just as it does to phrases. Indeed,
compositionality is often taken to be a  signal
for productivity \cite{aronoff1976word}.  When deciding whether to
further decompose a word, asking whether the parts sum up to the whole
is often a good indicator. In the case of \word{questionably}
$\mapsto$ \word{question}$+$\word{able}$+$\word{ly}, the
compositional meaning is
``in a manner that could be
questioned'', which corresponds to the meaning of the word. Contrast this with the
word \word{unquiet}, which means ``restless'', rather than ``not
quiet'' and the compound \word{blackmail}, which does not refer
to a letter written in black ink.

The model we will describe in \secref{joint} is a \emph{joint model of both
semantic coherence and segmentation}; that is, an analysis is judged
not only by character-level features, but also by the degree to which
the word is semantically compositional. Implicit in such a treatment
is the desire to only segment a word if the segmentation is derived
from a productive process. While most prior work on morphological segmentation
has not explicitly modeled productivity,\footnote{Note that segmenters such as {\sc Morfessor} utilize the principle of minimum description length, which implicitly encodes productivity, in order to guide segmentation.}
we believe, from a computational modeling perspective, segmenting only
productive affixes is preferable. This is analogous to the modeling
of phrase compositionality in embedding models, where 
it can be better to
not further decompose
noncompositional multiword units like named entities and
idiomatic expressions; see, e.g.,
\newcite{mikolov13phrases}, \newcite{wang14jointly}, \newcite{yin15discriminative}, \newcite{yaghoobzadeh15typing}, and \newcite{hashimoto16phrase}.\footnote{As
  a reviewer points out, productivity of an affix and
  semantic coherence of the words formed from it are not
  perfectly aligned. Nonproductive affixes can produce
  semantically coherent words, e.g., 
\word{warm}$\mapsto$\word{warm}$+$\word{th}. Productive
affixes can produce semantically incoherent words, e.g., 
\word{canny}$\mapsto$\word{un}$+$\word{canny}. Again, this
is analogous to multiword units. However, there is a strong
correlation and our experiments show that relying
on it gives good results.}

In this paper, we refer to the semantic aspect of the model
either as \emph{semantic synthesis} or as \emph{coherence}.
These are two ways of looking at semantics that are
related as follows.  If the synthesis (i.e., composition) of
the meaning of the derived form from the meaning of its
parts is a regular application of the linguistic rules of
derivation, then the meaning so constructed is coherent.
These are the cases where a joint model is expected to be
beneficial for both segmentation and interpretation.

\section{A Joint Model}\seclabel{joint}
From an NLP perspective, canonical segmentation
\cite{naradowsky2009improving,cotterell2016canonical} is the task
that seeks to algorithmically decompose a word into its {\em canonical} sequence of
morphemes. It is a version of morphological segmentation that requires
the learner to handle orthographic changes that take place during word formation.
We believe this is a more natural formulation of morphological
analysis---especially for the processing of derivational morphology---as it draws heavily on linguistic notions (see \secref{deriv-morph}).

The main innovation we present  is the augmentation of canonical segmentation
to take into account semantic coherence and productivity. Consider the  word  
\word{hypercuriosity}
and its canonical segmentation
\word{hyper}$+$\word{curious}$+$\word{ity}; this canonical segmentation
seeks to decompose the word into its constituent morphemes
and account for  orthographic changes. This amounts to a
{\em structural} decomposition of the word, i.e., how do we break up the
string of characters into chunks? This is similar to the decomposition of a
sentence into a parse tree. However, it is also natural to consider
the {\em semantic} compositionality of a word, i.e., how is the
meaning of the word synthesized from the meaning of the
individual morphemes?

We consider both of these questions together in a single model,
where we would like to place high probability
on canonical segmentations that are also semantically coherent.
Returning to
\word{hypercuriosity},
we could further decompose it into 
\word{hyper}$+$\word{cure}$+$\word{ous}$+$\word{ity}
in analogy to, say, \word{vice} $\mapsto$ \word{vicious}.
Nothing about the surface form of {\em curious} alone gives us a strong cue
that we should rule out the segmentation
\word{cure}$+$\word{ous}.
Turning to distributional semantics, however, it is the case that the
contexts in which \word{curious} occurs are quite different from those
in which \word{cure} occurs. This gives us a strong cue which
segmentation is correct.

%
%
%
%
%

\begin{figure}
  \includegraphics[width=0.49\textwidth]{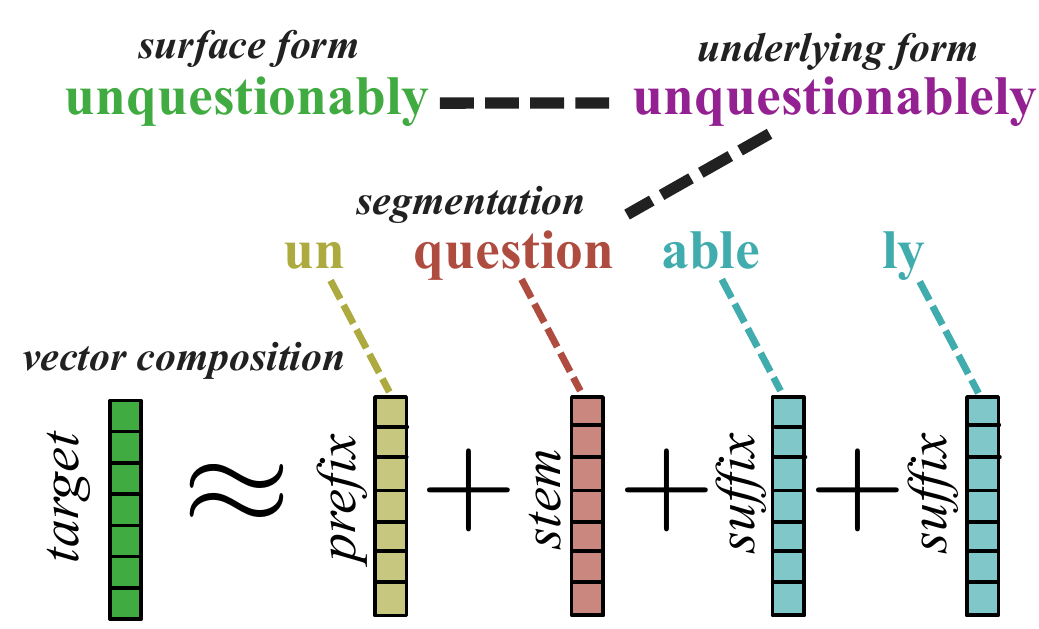}
  \caption{A depiction of the joint model that makes the relation between
    the three factors and the observed surface form explicit. We show a simple additive
    model of composition for ease of explication.
  \label{fig:joint}}
\end{figure}

Formally, given a word string $w \in
\Sigma^*$, where $\Sigma$ is a discrete alphabet of
characters 
(in English
this could be as simple as the 26 letter lowercase
alphabet), 
and a
word vector $v \in V$, where $V$ is a set of low-dimensional
word embeddings,
we define the model as:
\begin{align}\eqlabel{model}
  p(\vv, s, l, u \mid  &w)\nonumber\\
& = \frac{1}{Z_{\vtheta}(w)}\exp\left(-\frac{1}{2\sigma^2}||\vv -
  \CC(s, l) ||_2^2 \right.\nonumber \\
  &\left.+ \vf(s, l, u)^{\top} \veta + \vg(u, w)^{\top}\vomega \right).
\end{align}
This model is composed of three factors:
a composition factor, a segmentation factor and a transduction factor.
The parameters of the model are
$\vtheta = \{\vbeta, \veta, \vomega\}$,
the function $\mathcal{C}_{\vbeta}$ composes morpheme vectors together,
$s$ is the segmentation, $l$ is the labeling of the
segments, $u$ is the underlying representation and $Z_{\vtheta}(w)$ is the partition function.
Note that the conditional distribution $p(\vv \mid s, l, u, w)$ is Gaussian distributed by construction.
A visualization of our model is found in \figref{joint}.
This model is a conditional random field (CRF) that is mixed, i.e., it
is defined over both discrete and continuous random variables
\cite{koller2009probabilistic}.  We restrict the range of $u$ to
be a subset of $\Sigma^{|w|+k}$, where $k$ is an insertion limit \cite{dreyer2011non}. In this work,
we take $k=5$. Explicitly, the partition function is defined as
\begin{align}
  Z_{\vtheta}(w) &= \int \sum_{l', s', u'}\exp\left(-\frac{1}{2\sigma^2}||\vv' -
  \CC(s', l')||_2^2  \right.\nonumber \\ &+ \left. \vf(s', l', u')^{\top} \veta + \vg(u', w)^{\top}\vomega \right) d\vv',
\end{align}
which is guaranteed to be finite.\footnote{Since we have capped the insertion limit,
  we have a finite number of values that $u$ can take  for any $w$. Thus, it follows
  that we have a finite number of canonical segmentations $s$. Hence
  we take a finite number of Gaussian integrals. These integrals all converge since
  we have fixed the covariance matrix as $\sigma^2I$, which is positive definite.}

A CRF is simply the globally renormalized product of several
non-negative factors \cite{sutton2006introduction}. 
Our model is composed of three: transduction, segmentation and
composition factors---we describe each in turn.


\subsection{Transduction Factor}
The first factor we consider is the transduction factor:
$\exp\left(\vg(u, w)^{\top}\vomega\right)$, which scores a {\em
  surface representation} (SR) $w$, the character string observed in
raw text, and an {\em underlying representation} (UR), a character
string with orthographic processes
reversed. The aim of this factor is to place high weight on good
pairs, e.g., the pair ($w$=\word{questionably},$u$=\word{questionablely}), so we can
accurately restore character-level changes.

We encode this portion of the model as a weighted finite-state
machine for ease of computation. This factor generalizes 
probabilistic edit distance \cite{ristad1998learning} by looking at
additional input and output context; see
 \newcite{cotterell-peng-eisner-2014} for details.
As mentioned above and in contrast to
\newcite{cotterell-peng-eisner-2014}, we bound the insertion limit in
the edit distance model.\footnote{As our transduction model is an unnormalized factor in a
  CRF, we do not require the local normalization discussed in
  \newcite{cotterell-peng-eisner-2014}---a weight on an edge may
  be any non-negative real number since we will renormalize later. The
  underlying model, however, remains the same.} Computing the score between two strings $u$ and $w$ requires a dynamic program
that runs in $\mathcal{O}(|u|\cdot |w|)$. This is a generalization of the forward
algorithm for Hidden Markov Models (HMMs) \cite{rabiner1989tutorial}. 

We employ standard feature templates for the task that look at features
of edit operations, e.g., substitute $i$ for $y$, in
varying context granularities. See
\newcite{cotterell2016canonical} for details.
Recent work has also explored weighting of WFST arcs
with scores computed by LSTMs
\cite{hochreiter1997long}, obviating the need for human
selection of feature templates \cite{rastogi2016}.

\subsection{Segmentation Factor}
The second factor is the segmentation factor: $\exp\left(\vf(s,
l,u)^{\top} \veta\right)$.  The goal of this factor is to score a
segmentation $s$ of a UR $u$. In our example, it scores the input-output pair ($u$$=$\word{questionablely},
$s$$=$\word{question}$+$\word{able}$+$\word{ly}). It additionally scores a
labeling of the segmentation. Our label set in this work is
$L = \{\text{stem}, \text{prefix}, \text{suffix}\}$.  The proper
labeling of the segmentation above is $l$$=$\word{question}{\em
  :stem}$+$\word{able}{\em :suffix}$+$\word{ly}{\em :suffix}.
The labeling is critical for our composition
functions $\mathcal{C}_{\vbeta}$ \cite{cotterell2015labeled}:
which vectors are used depends on the label given
to the segment; e.g., the vectors of the prefix ``post'' and the stem
``post'' are different.


We can view this factor as an unnormalized first-order semi-CRF
\cite{sarawagi2004semi}. Computation of the factor  again requires
dynamic programming.  The algorithm is a different
generalization of the forward algorithm for HMMs, one that extends it to the semi-Markov case.
This algorithm runs in $\mathcal{O}({|u|^2\cdot |L|^2})$. 

\spacesavingparagraph{Features.}
We again use standard feature templates for the task. We create
atomic indicator features for the individual segments. We then conjoin
the atomic features with left and right context features as well as
the label to create more complex feature templates. We also include
transition features that fire on pairs of sequential labels. See
\newcite{cotterell2015labeled} for  details.
Recent work has also showed that a neural parameterization
can remove the need for manual feature design \cite{kong2015segmental}.



\begin{table}
  \setlength{\extrarowheight}{5pt}
  \begin{tabular}{l|lll}
    model & \multicolumn{3}{l}{composition function} \\ \hline
    {stem} & $c$ & = & $\sum_{i=1}^N \mathbbm{1}_{l_i = \text{stem}} m^{l_i}_{s_i}$ \\
    {mult} & $c$ & = &$\bigodot_{i=1}^N m^{l_i}_{s_i}$ \\
    {add} & $c$ & = &$\sum_{i=1}^N m^{l_i}_{s_i}$ \\
    {wadd} & $c$ & = &$\sum_{i=1}^N \alpha_i 
m^{l_i}_{s_i}$ \\
    {fulladd} & $c$ & = &$\sum_{i=1}^N U_i 
m^{l_i}_{s_i}$ \\
    {LDS}  & $h_i$& =& $X h_{i-1} + U m^{l_i}_{s_i}$\\
    {RNN} & $h_i$ &=& $\tanh(X h_{i-1} + U m^{l_i}_{s_i})$
  \end{tabular}
  \caption{Composition models $\mathcal{C}_{\vbeta}(s, l)$
    used in this and prior work. The representation of
    the word is 
$h_{N}$ for the dynamic 
and $c$ for the non-dynamic models.
  Note that for the dynamic models $h_0$ is a learned parameter.\tablabel{composition-models}}
\end{table}

\subsection{Composition Factor}
The composition factor takes the form of an unnormalized multivariate Gaussian
density: $\exp\left(-\frac{1}{2\sigma^2}||v -
\mathcal{C}_{\vbeta}(s, l)||_2^2\right)$, where the mean is computed
by the (potentially non-linear) composition function 
(See \tabref{composition-models})
and the
covariance matrix $\sigma^2I$ is a diagonal matrix.  The goal of the composition
function $\mathcal{C}_{\vbeta}(s, l)$ is to stitch together {\em
  morpheme} embeddings to approximate the vector of the entire word.


The simplest form of the composition function 
  $\mathcal{C}_{\vbeta}(s, l)$ is \emph{add}, an additive model of the morphemes.
See \tabref{composition-models}:
each vector $\vm^{l_i}_{s_i}$ refers to a morpheme-specific,
label-dependent embedding. If $l_i = \text{stem}$, then
$s_i$ represents a stem morpheme. 
Given that our
segmentation is
canonical, an $s_i$ that is a stem generally itself is an entry in the lexicon and
 $v(s_i) \in V$. If
$v(s_i) \not\in V$, then we set $v(s_i)$ to 0.\footnote{This
is not changed in training, so all such $v(s_i)$ are 0 in
the final model. Clearly, this could be improved in future
work as a reviewer points out, e.g., by setting 
such $v(s_i)$ to an average of a suitable chosen set of
known word vectors.}  We optimize
over vectors with $l_i \in \{\text{prefix}, \text{suffix}\}$ as
they correspond to bound morphemes.

We also consider a more expressive composition model,
a recurrent neural network (RNN).
Let  $N$ be the number of segments. Then
$\mathcal{C}_{\vbeta}(s, l) = h_{N}$
where $h_i$ is a hidden vector, defined
by the  recursion:\footnote{We do not explore
  more complex RNNs, e.g., LSTMs \cite{hochreiter1997long} and GRUs \cite{cho2014properties}
  as words in our data have $\leq$7
  morphemes. 
These architectures make the learning
  of long distance dependencies easier, but
  are no more powerful than an Elman RNN, at least in theory. Note that
  perhaps if applied to languages with richer derivational morphology than English,
considering more complex neural architectures would make sense.}
 $ h_i = \tanh\left( X h_{i-1}+U m^{l_i}_{s_i} \right)$ \cite{elman1990finding}.
Again, we optimize the morpheme embeddings $m_{s_i}^{l_i}$ only when $l_i \neq \text{stem}$
along with the other parameters of the RNN, i.e., the matrices $U$ and $X$.

\section{Inference and Learning}\seclabel{inf-learn}
Exact inference  is intractable since 
we allow arbitrary segment-level features on the canonicalized
word forms $u$. Since the semi-CRF factor has features
  that fire on substrings, we would need a dynamic programming state for each substring of each of the
  exponentially many settings of $u$; this
breaks the dynamic program. We thus turn to approximate inference through
an importance sampling routine \cite{rubinstein2011simulation}.

\subsection{Inference by Importance Sampling}
\seclabel{impsampling}
Rather than
considering all underlying orthographic forms $u$ and segmentations
$s$, we sample from a tractable proposal distribution $q$---a distribution
over canonical segmentations. In the
following equations we omit the dependence on $w$ for
notational brevity and define $\vh(l, s, u) = \vf(s,l,u) + \vg(u,w)$.
Crucially, the partition function $Z_{\vtheta}(w)$ is {\em not} a function
of parameter subvector $\vbeta$ and its gradient with respect
to $\vbeta$ is 0.\footnote{The subvector $\vbeta$ is responsible
for computing only the {\em mean} of the Gaussian factor and thus has no impact
on its normalization coefficient \cite{murphy2012machine}.} Recall that
computing the gradient of the log-partition function is equivalent
to the problem of marginal inference \cite{wainwright2008graphical}.
We derive our estimator as follows:
  \begin{flalign}
    \nabla_{\vtheta} \log Z
    &= \Expect_{(l, s,u) \sim p} \left[ \vh(l, s,u)\right] & \\
    &= \sum_{l, s, u} p(l, s, u) \vh(l, s,u) & \\
    &= \sum_{l, s, u} \frac{q(l, s, u)}{q(l, s, u)} p(l, s, u) \vh(l, s,u) & \\
    &= \Expect_{(l, s, u) \sim q}\left[ \frac{p(l, s, u)}{q(l, s, u)} \vh(l, s, u) \right],
  \end{flalign}
  \noindent
where we have omitted the dependence on $w$ (which we condition on) and $v$ (which we marginalize out).
So long as $q$ has support everywhere $p$ does (i.e., $p(l, s, u) > 0
\Rightarrow q(l, s, u) > 0$), the estimate is unbiased.
Unfortunately, we can
only efficiently compute $p(l, s, u)$
up to a constant factor, $p(l, s, u) = \bar{p}(l, s, u) / Z'_{\vtheta}(w)$. Thus, we use the
\emph{indirect importance sampling estimator},
\begin{eqnarray}
\label{indirect-IS}
\frac{1}{\sum_{i=1}^M  w^{(i)}} \sum_{i=1}^M  w^{(i)} \vh(l^{(i)}, s^{(i)}, u^{(i)}),
\end{eqnarray}
\noindent where $(l^{(1)}\!,\! s^{(1)}\!,\! u^{(1)}) \ldots (l^{(M)}\!,\! s^{(M)}\!,\! u^{(M)})$
$\stackrel{\text{i.i.d.}}{\sim} q$
and importance weights $w^{(i)}$ are defined as:
\begin{equation}\eqlabel{importance-weights}
  w^{(i)} = \frac{\bar{p}(l^{(i)}, s^{(i)}, u^{(i)})}{q(l^{(i)}, s^{(i)}, u^{(i)})}.
\end{equation}
This
indirect estimator is biased, but consistent.\footnote{Informally,
  the indirect importance sampling estimate converges to the \emph{true}
  expectation as $M \rightarrow \infty$ (the definition of statistical consistency).}

\spacesavingparagraph{Proposal Distribution.}
The success of importance sampling depends  on the choice of
a ``good'' proposal distribution, i.e., one that ideally is
close to $p$.  Since we are fully supervised at
training time, we have the option of training locally normalized
distributions for the individual components. Concretely, we train {\em
  two} proposal distributions $q_1(u \mid w)$ and $q_2(l, s \mid u)$ that
take the form of a WFST and a semi-CRF, respectively, using 
features identical to the joint model. Each of these
distributions is tractable---we can compute the marginals with dynamic
programming and thus sample efficiently. To draw samples $(l, s, u) \sim
q$, we sample sequentially from $q_1$ and then $q_2$, conditioned on the output of
$q_1$.

\subsection{Learning}
We optimize the log-likelihood of the model using \software{AdaGrad}
\cite{duchi2011adaptive}, which is SGD with a special per-parameter learning rate. 
The full gradient of the objective for one training example is:
\begin{align}
\nabla_{\vtheta} \log  p(\vv, s,l,u \mid w&) = \vf(s, l, u)^{\top} + \vg(u, w)^{\top}  \nonumber \\
 & -\frac{1}{\sigma^2}(\vv - \CC(s, l))\nabla_{\vtheta}\CC(s, l) \nonumber \\ &  - \nabla_{\vtheta} \log Z_{\vtheta}(w), 
\end{align}
where we use the importance sampling algorithm described in
\secref{impsampling} to approximate the gradient of the log-partition
function, following \newcite{bengio2003quick}. Note that $\nabla_{\vtheta}\CC(s, l)$ depends on the
composition function used. In the most complicated case when $\CC$ is
a RNN, we can compute $\nabla_{\vbeta}\CC(s, l)$ efficiently with
backpropagation through time \cite{werbos1990backpropagation}. We take
$M=10$ importance samples; using so few samples can lead to a poor
estimate of the gradient, but for our application it suffices.  We employ $L_2$ regularization.

\subsection{Decoding}
Decoding the model is also intractable. To approximate the solution,
we again employ importance sampling. We take $M=$10,000 importance
samples and select the highest weighted sample.

\begin{table*}
  \setlength{\tabcolsep}{4pt}
  \centering
  \begin{tabular}{ll||lll|lll}
    && \multicolumn{3}{c|}{{dev}} & \multicolumn{3}{c}{{test}} \\
    &Model & Acc & $F_1$ & Edit & Acc & $F_1$ & Edit \\ \hline\hline
\multirow{5}{*}{\rotatebox[origin=c]{90}{{EN}}}    &Semi-CRF (Baseline)                             & 0.55 \sss{(.018)} & 0.75 \sss{(.014)} & 0.80 \sss{(.043)} & 0.54 \sss{(.018)} & 0.75 \sss{(.014)} & 0.78 \sss{(.034)} \\
    &Joint (Baseline)                                & 0.77 \sss{(.011)} & 0.87 \sss{(.007)} & 0.41 \sss{(.029)} & 0.77 \sss{(.013)} & 0.87 \sss{(.007)} & 0.43 \sss{(.029)} \\
     & Joint + Vec (This Work)                         & 0.83 \sss{(.014)} & 0.91 \sss{(.008)} & 0.31 \sss{(.019)}  & 0.82 \sss{(.020)} & 0.90 \sss{(.011)} & 0.32 \sss{(.038)} \\
    &Joint + UR (Oracle)                    & 0.94 \sss{(.015)} & 0.96 \sss{(.009)} & 0.07 \sss{(.016)} & 0.94 \sss{(.011)} & 0.96 \sss{(.007)} & 0.07 \sss{(.011)} \\
    &Joint + UR + Vec (Oracle)             & 0.95 \sss{(.011)} & 0.97 \sss{(.007)} & 0.05 \sss{(.013)} & 0.95 \sss{(.023)} & 0.97 \sss{(.006)} & 0.05 \sss{(.025) } \\ \hline 
\multirow{5}{*}{\rotatebox[origin=c]{90}{{DE}}}    &Semi-CRF (Baseline)                                   & 0.39 \sss{(.062)} & 0.68 \sss{(.039)} & 1.15 \sss{(.230)}  & 0.39 \sss{(.058)}   & 0.68 \sss{(.042)} & 1.14 \sss{(.240)} \\
    &Joint (Baseline)                                      & 0.79 \sss{(.107)} & 0.88 \sss{(.069)} & 0.40 \sss{(.313)} & 0.79 \sss{(.099)}   & 0.87 \sss{(.063)} & 0.41 \sss{(.282)} \\
 &Joint + Vec (This Work)                         & 0.82 \sss{(.102)} & 0.90 \sss{(.067)} & 0.33 \sss{(.312)} & 0.82 \sss{(.096)}   & 0.90 \sss{(.061)} & 0.33 \sss{(.282)} \\
    &Joint + UR (Oracle)                                   & 0.86 \sss{(.108)} & 0.90 \sss{(.070)} & 0.25 \sss{(.288)} & 0.86 \sss{(.100)}   & 0.90 \sss{(.064)} & 0.25 \sss{(.268)} \\
    &Joint + UR + Vec (Oracle)                             & 0.87 \sss{(.106)} & 0.92 \sss{(.069)} & 0.20 \sss{(.285)} & 0.88 \sss{(.096)}   & 0.93 \sss{(.062)} & 0.19 \sss{(.263)} \\
  \end{tabular}
  \caption{Results for the canonical morphological segmentation task on English and German. Standard deviation is given in parentheses. We compare against two baselines that do not
    make use of semantic vectors: (i) ``Semi-CRF
    (baseline)'', a semi-CRF
    that {\em cannot} account for orthographic changes and
    (ii) ``Joint (Baseline)'', a version of our joint model without vectors. We also
    compare against an oracle version with
    access to  gold URs (``Joint + UR (Oracle)'', ``Joint +
    UR + Vec (Oracle)''), revealing that the toughest
    part of the canonical segmentation task is reversing the orthographic changes.\tablabel{canseg-results}}
\end{table*}

\section{Related Work}\seclabel{related-work}

%
%
%

The idea that vector semantics is useful for morphological segmentation is not
new. Count vectors \cite{salton71smart,turney2010frequency} 
have been shown to be
beneficial in the unsupervised induction of morphology
\cite{schone2000knowledge,schone2001knowledge}. 
Embeddings
were shown to act similarly
\cite{soricut2015unsupervised}. Our method differs from this
line of research in
two key ways. (i) We present a \emph{probabilistic} model of the process of
synthesizing the word's meaning from the meaning of its
 morphemes. Prior work was either not probabilistic or did
 not explicitly model morphemes.
(ii) Our method is supervised and focuses on
derivation.  \newcite{schone2000knowledge} and
\newcite{soricut2015unsupervised}, being fully unsupervised, do not
distinguish between inflection and derivation and \newcite{schone2001knowledge} focus on inflection.
More recently, \newcite{narasimhan2015unsupervised} look
at the unsupervised induction of ``morphological chains'' with
semantic vectors as a crucial feature. Their goal is to jointly
figure out an ordering of word formation and a morphological segmentation, e.g.,
\word{play}$\mapsto$\word{playful}$\mapsto$\word{playfulness}.
While it is a rich model like ours, theirs
differs in that it is unsupervised
and uses vectors as features, rather than explicitly treating
vector composition. All of the above work focuses
on {\em surface segmentation} and not {\em canonical segmentation}, as we do.

A related line of work that has different goals concerns morphological
generation.  Two recent papers that address this problem using deep
learning are \newcite{faruqui16lexgen} and \newcite{faruqui16seq2seq}.  In an older
line of work, \newcite{yarowsky2000minimally} and
\newcite{wicentowski2002modeling} exploit log frequency ratios of
inflectionally related forms to tease apart that, e.g., the past tense
of \word{sing} is not \word{singed}, but instead \word{sang}. Related
work by \newcite{dreyer2011discovering} uses a Dirichlet process to
model a corpus as a ``mixture of a paradigm'', allowing for the
semi-supervised incorporation of distributional semantics into a
structured model of inflectional paradigm completion.


\eat{An older line of work considers the role of distributional
signatures in lemmatization and inflection generation
\cite{yarowsky2000minimally,wicentowski2002modeling}.  The key
observation is that the log frequency ratio of inflectionally related
forms remains relatively constant across a language; this helps tease
apart that the past tense form of \word{sing} is not \word{singed},
the past tense of \word{singe}, but instead the irregular
\word{sang}. Related work by \newcite{dreyer2011discovering} uses a Dirichlet
process to model a corpus as a ``mixture of a paradigm'', allowing for
the semi-supervised incorporation of distributional semantics into a
structured model of inflectional paradigm completion.}

Our work is also related to recent attempts to integrate
morphological knowledge into general embedding models. For example,
\newcite{botha2014compositional} train a log-bilinear language model
that models the composition of morphological structure. Likewise,
\newcite{luong2013better} train a recursive neural network
\cite{goller1996learning} over a heuristically derived tree structure
to learn morphological composition over continuous vectors. Our work
is different in that we learn a joint model of
segmentation and composition.  Moreover, supervised morphological
analysis can drastically outperform unsupervised analysis
\cite{ruokolainena2013supervised}.

Early work by \newcite{kay1977morphological} can be interpreted as
finite-state canonical segmentation, but it neither addresses nor
experimentally evaluates the question of joint modeling of
morphological analysis and semantic synthesis.  Moreover, we may view
canonicalization as an orthographic analogue to phonology. On this
interpretation, the finite-state systems of \newcite{kaplan94rule},
which computationally apply SPE-style phonological rules
\cite{ch68}, may be run backwards to get canonical underlying forms.


\section{Experiments and Results}\seclabel{experiments}

We conduct experiments on English and German derivational
morphology. We analyze our joint model's ability to segment words into
their canonical morphemes as well as its ability to compositionally
derive vectors for new words. Finally, we explore the relationship
between distributional semantics and morphological productivity.

For English, we use the pretrained vectors of
\newcite{levy2014dependency} for all experiments. For German, we train
word2vec skip-gram vectors on the German Wikipedia. We first describe
our English dataset, the subset of the English portion of the CELEX
lexical database \cite{baayen1993celex} that was selected by
\newcite{LazaridouMZB13}; the dataset contains 10,000 forms. This allows for comparison with previously
proposed methods.  We make two modifications.
(i) \newcite{LazaridouMZB13} make the \emph{two-morpheme assumption}: every
word is
composed of exactly two morphemes. In general, this is not true, so we
further segment all complex words in the corpus. For example,
\word{friendless}$+$\word{ness} is further segmented into
\word{friend}$+$\word{less}$+$\word{ness}. 
To nevertheless allow for
fair comparison, we provide versions of our experiments with and
without the two-morpheme assumption where
appropriate.
(ii) \newcite{LazaridouMZB13} only provide a single train/test split.
As we require a held-out development set for
hyperparameter tuning, we randomly allocate a portion of the training
data to select the hyperparameters and then retrain the model using
these parameters on the original train split. We also
report 10-fold cross validation results in addition to
Lazaridou et al.'s   train/test
split.

Our German dataset is taken from \newcite{zeller2013derivbase} and is
described in \newcite{cotterell2016canonical}. It, again, consists
of 10,000 derivational forms. We report results on 10-fold
cross validation.

\subsection{Experiment 1: Canonical Segmentation}\seclabel{exp-can-seg}
For our first experiment, we test whether jointly modeling the continuous
representations allows us to segment words more accurately. We
assume that we are given an embedding for the target word. We estimate
the model $p(v, s, l, u \mid w)$ as described in \secref{inf-learn}
with $L_2$ regularization $\lambda||\vtheta||_2^2$. To evaluate, we decode the distribution $p(s, l, u \mid v, w)$.
We perform approximate MAP inference with importance sampling---taking
the sample with the highest score. In these experiments, we use the
RNN with the dependency vectors, the combination of which performs best
on vector approximation in \secref{exp-vec-approx}.

We follow the experimental design of \newcite{cotterell2016canonical}. We
compare against two baselines (marked ``Baseline'' in \tabref{canseg-results}): (i) a ``Semi-CRF'' segmenter that cannot
account for  orthographic changes and (ii) the full ``Joint'' model of
\newcite{cotterell2016canonical}.\footnote{i.e., a model
{\em without} the Gaussian factor that scores vectors.} We
additionally consider an ``Oracle'' setting, where we give the model
the gold underlying orthographic form (``UR'') at both training and test time.
This gives us insight into the performance of the transduction factor
of our model, i.e., how much could we benefit from a richer model.

%
%

Our hyperparameters are (i) the regularization coefficient
$\lambda$ and (ii) $\sigma^2$, the variance of the Gaussian
factor. We use grid search to tune them: $\lambda \in
\{0.0, 10^{1}, 10^{2}, 10^{3}, 10^{4}, 10^{5}\}$, $\sigma^2 \in
\{0.25, 0.5, 0.75, 1.0\}$.

\def\oraclejointspace{0.00cm}
\begin{table}
  \centering
  \begin{adjustbox}{width=.475\textwidth}
    \begin{tabular}{ll||cc|cc|cc||cc}
      &    & \multicolumn{6}{c||}{{EN}} & \multicolumn{2}{c}{{DE}} \\ \cline{3-10}
&    & \multicolumn{2}{c|}{{BOW2}} & \multicolumn{2}{c|}{{BOW5}} & \multicolumn{2}{c||}{{DEPs}} & \multicolumn{2}{c}{{SG}} \\
&    & dev & test & dev & test & dev & test & dev & test \\ \hline\hline
    \multirow{4}{*}{\rotatebox[origin=c]{90}{{\footnotesize
          oracle}}}&
     stem          & .403  & .402  & .374  & .376  & .422         & .422       & .400 & .405\\
&     add          & .635  & .635  & .541  & .542  & .787         & .785       & .712 & .711\\
&     LDS          & .660  & .660  & .566  & .568  & .806
     & .804       & {\bf .717} & {\bf .718} \\
&     RNN          & .660  & .660  & .565  & .567  & {\bf .807}   & {\bf .806} & .707 & .712   \\ \hline\hline
    \multirow{4}{*}{\rotatebox[origin=c]{90}{{\footnotesize
          joint}}}&
     stem           & .399  & .400  & .371  & .372  & .411        & .412  & .394 & .398  \\
&     add           & .625  & .625  & .524  & .525  & .782        & .781  &  .705 & .704  \\
&     LDS           & .648  & .648  & .547  & .547  & .799
     & .797  & {\bf .712} & { \bf .711}     \\
&     RNN           & .649  & .647  & .547  & .546  & {\bf .801}  & {\bf .799} & .706 &  .708 \\ \hline
    \multirow{2}{*}{\rotatebox[origin=c]{90}{{\footnotesize
          char}}}&
    GRU             & .586 & .585   & .452  & .452  & .769        & .768 & .675 & .667 \\
&    LSTM           & .586 & .586   & .455  & .455  & .768        & .767 & .677 & .666
  \end{tabular}
  \end{adjustbox}
  \caption{Vector approximation (measured by mean cosine
    similarity) both with (``oracle'') and without 
(``joint'', ``char'')
gold
    morphology.
Surprisingly, joint models are
    close in performance to models with gold morphology.
\tablabel{vec-approx-no-gold}}
\end{table}

\spacesavingparagraph{Metrics.}
We use three metrics to evaluate  segmentation accuracy. Note that
the evaluation of canonical segmentation is  hard since a system may return a
sequence of morphemes whose concatenation is not the same length as
the concatenation of the gold morphemes. This rules out
metrics for surface segmentation like border $F_1$ \cite{kurimo2010morpho},
which require the strings to be of the same length.

We now define the metrics. (i) \emph{Segmentation
  accuracy} measures whether every
single canonical morpheme in the returned sequence is correct. It
is  inflexible: closer answers are penalized the same as
more distant answers. 
 (ii)
\emph{Morpheme $F_1$} \cite{van1999memory} takes the predicted sequence of canonical
morphemes, turns it into a set,  computes precision and
recall in the standard way and based on that then computes $F_1$. This metric gives credit
if some of the canonical morphemes were correct. 
 (iii) \emph{Levenshtein distance}
 joins the canonical segments with a
special symbol \# into a single string and  computes
the Levenshtein distance between predicted and gold strings. 

\spacesavingparagraph{Discussion.}
Results 
in \tabref{canseg-results}
show  that jointly modeling semantic coherence
improves our ability to analyze words. For test,
our proposed joint model (``This Work'')
outperforms the baseline supervised canonical segmenter, which is
state-of-the-art for the task, by
.05 (resp.\ .03) on accuracy and .03 (resp.\ .03) on $F_1$
for English (resp.\ German).
We also find that when we give the
joint model an oracle UR the vectors generally help
less:
.01 (resp.\ .02) on accuracy and .01 (resp.\ .03) on $F_1$ for
English (resp.\ German).
 This indicates that the chief boon the vector composition
factor provides lies in selection of an appropriate UR. Moreover, the
up to .15  difference in English between systems with and without the oracle UR
suggests that reversing  orthographic changes is a
particularly difficult part of the task, at least for English.

\subsection{Experiment 2: Vector Approximation}\seclabel{exp-vec-approx}
We adopt the experimental design of
\newcite{LazaridouMZB13}. Its aim
is to approximate a vector of a derivationally complex word using
a learned model of composition. As \newcite{LazaridouMZB13} assume
a gold morphological analysis, we compare two settings: (i) oracle morphological analysis
and (ii) inferred morphological analysis. To the best of our knowledge,
(ii) is a novel experimental condition that no previous work
has 
 addressed.

%
%
%

We consider four composition models
(See \tabref{composition-models}). (i) \emph{stem}, using
just the stem vector. This baseline tells us what happens if
we make the incorrect assumption that derivation behaves
like inflection and is not meaning-changing.
(ii) \emph{add}, a purely additive
model. This is arguably the simplest way of combining 
the vectors of the morphemes.
 (iii) \emph{LDS}, a linear
dynamical system. This is arguably the simplest sequence
model.
(iv) A (simple) \emph{RNN}.  Recurrent neural networks
are currently the most widely used
nonlinear
sequence model and simple RNNs are the simplest such models.

Part of the motivation for considering a
richer class of models lies in our removal of the two-morpheme
assumption. Indeed, it is unclear that the {\em wadd} and {\em fulladd} models \cite{mitchell2008vector}
are useful models in the general case of multi-morphemic words---the weights
are tied by {\em position}, i.e., the first
morpheme's vector (be it a prefix or stem)
is always multiplied by the same matrix.

\def\goldmorphspace{0.1cm}

\setlength\belowcaptionskip{-7.5pt}
\begin{table}[t]
  \begin{tabular}{cl||c@{\hspace{\goldmorphspace}}c@{\hspace{\goldmorphspace}}c|c@{\hspace{\goldmorphspace}}c@{\hspace{\goldmorphspace}}c}
    && all & HR & LR & { -less} & { in-} & { un-} \\ \cmidrule{1-8}\morecmidrules\cmidrule{1-8}
      \multirow{5}{*}{\rotatebox[origin=c]{90}{{\scriptsize Lazaridou}}}
    &  {stem}    & .47 & .52 & .32 & .22 & .39 & .33 \\
    & {mult}    & .39 & .43 & .28 & .23 & .34 & .33 \\
    & {dil.}    & .48 & .53 & .33 & .30 & .45 & .41 \\
    & {wadd}    & .50 & .55 & .38 & .24 & .40 & .34 \\
    & {fulladd} & .56 & .61 & .41 & .38 & .47 & .44 \\ \cmidrule{1-8}\morecmidrules\cmidrule{1-8}
    & {lexfunc} & .54 & .58 & .42 & .44 & .45 & .46 \\ 
    \multirow{7}{*}{\rotatebox[origin=c]{90}{{\scriptsize BOW2}}}
    & {stem}    & .43 & .44 & .38 & .32 & .43 & .51 \\
    & {add}     & .65 & .67 & .61 & .60 & .64 & .67 \\
    & {LDS}     & .67 & .69 & .62 & .61 & .66 & .67 \\
    & {RNN}     & .67 & .69 & .60 & .60 & .65 & .66 \\ \cline{2-8}
    & {c-GRU}   & .59 & .60 & .55 & .59 & .55 & .57 \\
    & {c-LSTM}  & .52 & .53 & .50 & .55 & .50 & .50 \\ \cmidrule{1-8}\morecmidrules\cmidrule{1-8}
    \multirow{6}{*}{\rotatebox[origin=c]{90}{{\scriptsize BOW5}}}
    & {stem}    & .40 & .43 & .33 & .27 & .37 & .46 \\
    & {add}     & .56 & .59 & .51 & .46 & .55 & .59 \\
    & {LDS}     & .58 & .61 & .51 & .48 & .57 & .60 \\
    & {RNN}     & .58 & .61 & .50 & .48 & .56 & .58 \\ \cline{2-8}
    & {c-GRU}   & .45 & .47 & .42 & .42 & .43 & .45 \\
    & {c-LSTM}  & .46 & .47 & .43 & .43 & .45 & .46 \\ \cmidrule{1-8}\morecmidrules\cmidrule{1-8}
    \multirow{6}{*}{\rotatebox[origin=c]{90}{{\scriptsize  DEPs}}}
    & {stem}    & .46 & .45 & .49 & .38 & .57 & .67 \\
    & {add}     & .79 & .79 & .77 & .78 & .80 & .80 \\
    & {LDS}     & .80 & .81 & {\bf .77} & {\bf .79} & {\bf .81} & {\bf .81} \\
    & {RNN}     & {\bf .81} & {\bf .82} & {\bf .77} & {\bf .79} & .80 & {\bf .81} \\ \cline{2-8}
    & {c-GRU}   & .75 & .76 & .72 & .78 & .74 & .75 \\
    & {c-LSTM}  & .75 & .76 & .71 & .77 & .72 &
    .73 
  \end{tabular}
  \caption{Vector approximation 
(measured by mean cosine similarity) 
with gold morphology
    on the train/test split of \newcite{LazaridouMZB13}. 
    HR/LR = high/low-relatedness words. See \newcite{LazaridouMZB13} for  details.
\tablabel{oracle-vector-approx}}
\end{table}

\spacesavingparagraph{Comparison with Lazaridou et al.}
To
compare with  \newcite{LazaridouMZB13}, we use
their exact train/test split. Those results are reported in
\tabref{oracle-vector-approx}. This dataset enforces that all words are
composed of exactly two morphemes.  Thus, a word like
\word{unquestionably} is segmented as
\word{un}$+$\word{questionably}, without further decomposition.  The
vectors employed by \newcite{LazaridouMZB13} are
high-dimensional count vectors derived from lemmatized and POS tagged
text with a before-and-after window of size 2. They then apply 
pointwise mutual information (PMI) weighting  and dimensionality
reduction by non-negative matrix factorization. In contrast, we employ
\software{word2vec}  \cite{mikolov2013efficient}, a model
that is also interpretable as
the factorization of
a PMI matrix \cite{levy2014neural}. We consider three 
\software{word2vec} models: two bag-of-word (BOW) models with
before-and-after windows of size 2 and 5 and
DEPs \cite{levy2014dependency},
a dependency-based model
whose context is derived from
dependency parses rather than BOW.

In general, the results indicate that the key  to better
vector approximation is not a richer model of composition, but
rather lies in the vectors themselves. We find that our best model, the
RNN, only marginally edges out the LDS. Additionally, 
looking at the ``all'' column and the DEPs vectors,
the simple
additive model is only $\leq$.02 lower than LDS.
In comparison, we
observe large differences between the vectors. The RNN+DEPs model 
is .23 better than
the BOW5 models (.81 vs.\ .58), .14 better than the BOW2
models 
(.81 vs.\ .67)
and .25
better than
Lazaridou et al.'s best model (.81 vs.\ .56). 
A wider context for BOW (5 instead of 2)
yields   worse results. This suggests  that syntactic information
or at least positional information  is necessary for
improved models of morpheme composition.
The test vectors are annotated for relatedness, which is a
proxy for semantic coherence.  HR (high-relatedness) words were judged
to be more compositional than LR (low-relatedness) words.

\spacesavingparagraph{Character-Level Neural Retrofitting.}
As a further strong baseline, we consider a retrofitting
\cite{faruqui2015retrofitting} approach based on character-level
recurrent neural networks. Recently, running a recurrent net over the
character stream has become a popular way of incorporating subword
information into a model---empirical gains have been observed in
a diverse set of NLP tasks: POS tagging
\cite{santos2014learning,LingDBTFAML15}, parsing
\cite{ballesterosimproved} and language modeling \cite{KimJSR16}.
To the best of our knowledge, character-level retrofitting is a novel approach.

Given a vector $v$ for a word form $w$, we seek a function
to minimize the following objective
\begin{equation}
  \frac{1}{2}||v - h_N||_2^2, 
\end{equation}
where $h_N$ is the final hidden state of a recurrent neural architecture, i.e.,
\begin{equation}
  h_i = \sigma(A h_{i-1} + B w_i),
\end{equation}
where $\sigma$ is a non-linearity and $w_i$ is the $i^\text{th}$
character in $w$, $h_{i-1}$ is the previous hidden state and $A$ and
$B$ are matrices. While we have defined the architecture for a vanilla
RNN, we experiment with two more advanced recurrent architectures:
GRUs \cite{cho2014learningLSTM} and LSTMs \cite{hochreiter1997long} as
well as deep variants
\cite{sutskever14seq2seq,gillick15bytes,firat16multiway}.
Importantly, this model has {\em no knowledge} of morphology---it can
only rely on representations it extracts from the characters. This
gives us a clear ablation on the benefit of adding structured
morphological knowledge.  We optimize the depth and the size of
the hidden units on development data using a coarse-grained grid
search. We found a depth of 2 and hidden units of size 100 (in both
LSTM and GRU) performed best. We trained all models for 100 iterations
of Adam \cite{kingma2014adam} with $L_2$ regularization with
regularization coefficient $0.01$.

\tabref{oracle-vector-approx} shows that
the two character-level models 
(``c-GRU'' and ``c-LSTM'')
perform much worse than our models.
This indicates that supervised morphological analysis produces
higher-quality vector representations than ``knowledge-poor''
character-level models. However, we note that these character-level
models have fewer parameters than our morpheme-level models---there
are many more morphemes in a languages than characters. 

\spacesavingparagraph{Oracle Morphology.}
In general, the two-morpheme assumption is incorrect. We
consider an expanded setting of
\newcite{LazaridouMZB13}'s task, in which we fully decompose the
word, e.g., \word{unquestionably}$\mapsto$\word{un}$+$\word{question}$+$\word{able}$+$\word{ly}.  These 
results are reported in \tabref{vec-approx-no-gold} (top
block, ``oracle''). We report
mean cosine similarity.
Standard deviations $s$ for 10-fold cross-validation (not
shown) are small
  ($\leq .012$) with two exceptions: $s=.044$ for the
  DEPs-joint-stem results (.411 and .412).

\begin{figure*}
\centering
  \includegraphics[width=1.0\textwidth]{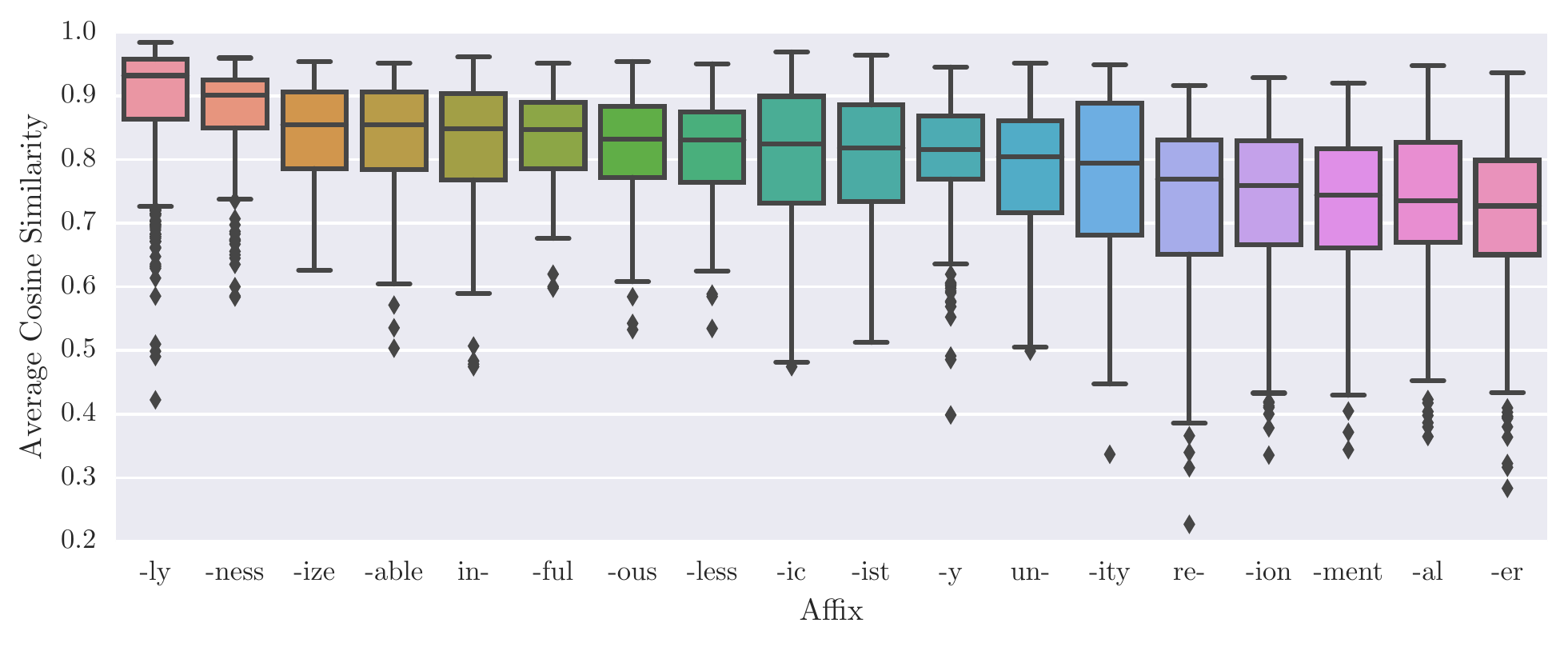}
  \caption{The boxplot breaks down the cosine similarity between the approximated
    vector and the target vector by affix (using gold morphology). We have ordered the affixes such that
    the better approximated vectors are on the left. 
  \figlabel{boxplot}}
\end{figure*}

The multi-morphemic results mirror those of
the bi-morphemic setting of \newcite{LazaridouMZB13}. 
(i) 
RNN+DEPs attains an
average cosine similarity of around .80 for English. Numbers for German
are lower, around .70.
(ii) The RNN
only marginally edges out LDS for English and is slightly
worse for German. Again, this is 
not    surprising as we are modeling short
    sequences. (iii) Certain embeddings
    lend themselves more naturally to derivational
    compositionality: 
BOW2
    is better than BOW5,
DEPs is the clear winner.

\eat{
\begin{align}
  \hat{v} &= \int v p(v \mid w) dv \\
  &\approx \frac{1}{M}\sum_{i=1}^M
  w^{(i)} \CC(l^{(i)}, s^{(i)}),
\end{align}
}

\spacesavingparagraph{Inferred Morphology.}
The final setting we consider is the vector approximation task without gold
morphology. In this case,  we rely on the full joint model $p(v, s, l, u \mid w)$.
At evaluation, we are interested in the marginal distribution $p(v \mid w) = \sum_{s, l, u}p(v, s, l, u \mid w)$.
We then use importance sampling to approximate the mean of this marginal distribution as the predicted embedding, i.e., 
\begin{align}
  \hat{v} &= \int v p(v \mid w) dv \\
  &\approx \frac{1}{\sum_{i=1}^M w^{(i)}}\sum_{i=1}^M
  w^{(i)} \CC(l^{(i)}, s^{(i)}), 
\end{align}
where $w^{(i)}$ are the importance weights defined in
\eqref{importance-weights} and $l^{(i)}$ and $s^{(i)}$ are the
$i^{\text{th}}$ sampled labeling and segmentation, respectively.

%
%
%
%
%

\spacesavingparagraph{Discussion.}
Surprisingly, \tabref{vec-approx-no-gold} (joint) shows that relying on the inferred morphology does not
drastically affect the results. Indeed, we are often within .01
of the result with gold morphology.
Our method can be viewed as a retrofitting procedure
\cite{faruqui2015retrofitting}, so this result is useful: it indicates that 
joint semantic synthesis and morphological analysis produces
high-quality vectors.

\subsection{Experiment 3: Derivational Productivity}\seclabel{exp-productivity}
We now delve into the relation between distributional
semantics and morphological productivity. The extent to which
jointly modeling semantics aids morphological analysis will
be determined by the inherent compositionality of the
words within the vector space. We break down our results on the vector approximation
task with gold morphology using the dependency vectors and the RNN composer in
\figref{boxplot} by selected affixes. 
We observe
a wide range of scores: the
most compositional ending \word{ly} gives rise to cosine
similarities that are  20 points higher than those of the
least compositional \word{er}.


On the left end of \figref{boxplot} we see extremely productive
suffixes. The affix \word{ize} is used productively with relatively
obscure words in the sciences, e.g., \word{Rao-Blackwellize}.
Likewise, the affix \word{ness} can be applied to almost any adjective
without restriction, e.g., \word{Poissonness} `degree to which data
have a Poisson distribution'.  On the right end, we find \word{-ment},
\word{-er} and \word{re-}.  The affix \word{-ment} is borderline
productive \cite{bauer1983english}---modern English tends to form
novel nominalizations with \word{ness} or \word{ity}. More interesting
are \word{re-} and \mbox{\word{er}}, both of which are very
productive in English. For \word{er}, many of the words
bringing down the average are simply non-compositional. For
example, \word{homer} `homerun in baseball' is not derived
from \word{home}$+$\word{er}---this is an error in data. We
also see examples like \word{cutter}. It has a 
compositional reading (e.g., ``box cutter''), but also
frequently occurs in the non-compositional meaning `type of
boat'.  Finally, proper nouns like \word{Homer} and
\word{Turner} end in \word{er} and in our experiments we
computed vectors for lowercased words.  The affix \word{re-}
similarly has a large number of non-compositional cases,
e.g., \word{remove}, \word{relocate}, \word{remark}. Indeed,
to get the compositional reading of \word{remove}, the first
syllable (rather than the second) is typically stressed to
emphasize the prefix.

We finally note several limitations of this experiment. (i)
The ability of our models---even the recurrent neural network---to model
transformations between vectors is limited. 
(ii) Our vectors are far from perfect;  e.g., sparseness in the training data affects quality
and some of the words in our corpus are rare.
(iii)
Semantic coherence is not the only
criterion for productivity. 
An example is \word{-th} in English. As noted earlier, it is
compositional in a word like \word{warmth},
but it cannot be used to form new
words.


\section{Conclusion}\seclabel{conclusion}
We have presented a model of the semantics and structure of
derivationally complex words. To the best of our knowledge,
this is the first attempt to jointly consider, within a
single model, (i) the morphological decomposition of the word form and
(ii) the semantic coherence of the resulting analysis. We
found that directly modeling coherence increases
segmentation accuracy, improving over a strong
baseline. Also, our models show state-of-the-art
performance on the derivational vector approximation task
introduced by \newcite{LazaridouMZB13}.

Future work will focus on the extension of the method to more complex
instances of derivational morphology, e.g., compounding and
reduplication, and on the extension to additional
languages. We also
plan to explore the relation between derivation
and distributional semantics in greater detail.

\enote{hs}{URL for datasets  here in final non-anonymous version}

\spacesavingparagraph{Acknowledgments.}
The first author was supported by a DAAD Long-Term Research
Grant and an NDSEG fellowship and
the second  by 
a
Volkswagenstiftung Opus Magnum grant. We would
also like to thank action editor Regina Barzilay
for suggesting several changes we incorporated
into the work and the three anonymous reviewers. 

\bibliographystyle{acl2012}
\bibliography{compositional-vectors}

\end{document}